\documentclass[sigconf]{acmart}
\usepackage{adjustbox}
\usepackage{graphics}
\usepackage{amsmath}
\usepackage{enumitem}
\usepackage[ruled,linesnumbered,vlined]{algorithm2e}
\usepackage{algorithmicx}
\usepackage{algpseudocode}

\usepackage{multirow}
\usepackage[normalem]{ulem}
\useunder{\uline}{\ul}{}

\AtBeginDocument{%
  \providecommand\BibTeX{{%
    \normalfont B\kern-0.5em{\scshape i\kern-0.25em b}\kern-0.8em\TeX}}}

\copyrightyear{2023}
\acmYear{2023}
\setcopyright{acmcopyright}\acmConference[WSDM '23]{Proceedings of the Sixteenth ACM International Conference on Web Search and Data Mining}{February 27-March 3, 2023}{Singapore, Singapore}
\acmBooktitle{Proceedings of the Sixteenth ACM International Conference on Web Search and Data Mining (WSDM '23), February 27-March 3, 2023, Singapore, Singapore}
\acmPrice{15.00}
\acmDOI{10.1145/3539597.3570455}
\acmISBN{978-1-4503-9407-9/23/02}

\acmConference[WSDM '23]{WSDM '23: ACM International Conference on Web Search and Data Mining}{February 27--March 3, 2023}{Singapore}
\acmBooktitle{WSDM '23: ACM International Conference on Web Search and Data Mining,
February 27--March 3, 2023, Singapore}

\begin{document}

\title[Self-Supervised Graph Structure Refinement for Graph Neural Networks]{Self-Supervised Graph Structure Refinement \\
for Graph Neural Networks}


\author{Jianan Zhao}\authornote{The first two authors contributed equally. Jianan Zhao is currently a student in Universit\'e de Montr\'eal.}
\affiliation{%
  \institution{University of Notre Dame}
  \city{Notre Dame, IN}
  \country{USA}}
\email{andy.zhaoja@gmail.com}

\author{Qianlong Wen}
\authornotemark[1]
\affiliation{%
  \institution{University of Notre Dame}
  \city{Notre Dame, IN}
  \country{USA}}
\email{qwen@nd.edu}

\author{Mingxuan Ju}
\affiliation{%
  \institution{University of Notre Dame}
  \city{Notre Dame, IN}
  \country{USA}}
\email{mju2@nd.edu}

\author{Chuxu Zhang}
\authornotemark[2]
\affiliation{%
  \institution{Brandeis University}
  \city{Waltham, MA}
  \country{USA}
}
\email{chuxuzhang@brandeis.edu}

\author{Yanfang Ye}
\authornote{Corresponding authors.}
\affiliation{%
  \institution{University of Notre Dame}
  \city{Notre Dame, IN}
  \country{USA}
}
\email{yye7@nd.edu}

\begin{abstract}
Graph structure learning (GSL), which aims to learn the adjacency matrix for graph neural networks (GNNs), has shown great potential in boosting the performance of GNNs. Most existing GSL works apply a joint learning framework where the estimated adjacency matrix and GNN parameters are optimized for downstream tasks. However, as GSL is essentially a link prediction task, whose goal may largely differ from the goal of the downstream task. The inconsistency of these two goals limits the GSL methods to learn the potential optimal graph structure. Moreover, the joint learning framework suffers from scalability issues in terms of time and space during the process of estimation and optimization of the adjacency matrix. To mitigate these issues, we propose a graph structure refinement (GSR) framework with a pretrain-finetune pipeline. Specifically, The pre-training phase aims to comprehensively estimate the underlying graph structure by a multi-view contrastive learning framework with both intra- and inter-view link prediction tasks. Then, the graph structure is refined by adding and removing edges according to the edge probabilities estimated by the pre-trained model. Finally, the fine-tuning GNN is initialized by the pre-trained model and optimized toward downstream tasks. With the refined graph structure remaining static in the fine-tuning space, GSR avoids estimating and optimizing graph structure in the fine-tuning phase which enjoys great scalability and efficiency. Moreover, the fine-tuning GNN is boosted by both migrating knowledge and refining graphs. Extensive experiments are conducted to evaluate the effectiveness (best performance on six benchmark datasets), efficiency, and scalability (13.8$\times$ faster using 32.8\% GPU memory compared to the best GSL baseline on Cora) of the proposed model. 
\end{abstract}
\vspace{-5pt}


\maketitle
\vspace{-5pt}
\section{Introduction}
\label{sec: intro}

With the proliferation of interaction systems such as recommendation and social networks,
various graph neural network (GNN) models~\cite{GCN,ChebNet,GAT,GIN,zhang2019heterogeneous} are proposed and have achieved remarkable improvements on a series of graph applications such as node classification~\cite{GCN,GAT}, graph classification~\cite{graph_cla_p1,graph_cla_p2,GIN}, recommendation~\cite{PinSAGE,GraphRec,NGCF}, and anomaly detection~\cite{qian2021distilling,qian2022malicious,qian2021adapting,qian2022rep2vec}. Most GNNs learn node representation by a message-passing scheme where node representation of each node is obtained by aggregating information from its neighbors.

Despite the success of GNNs, this message-passing scheme fundamentally relies on the assumption that the graph structure is reliable for downstream tasks. However, this assumption often fails to hold in real-world graphs~\cite{network_science}. This is partially because graphs are usually extracted from real-world interaction systems where sparseness and noise commonly exist~\cite{LDS,RobustGraphFromNoisyData,RobustStructuralNoise}. For example, in a user-item graph, it is well accepted that users may interact with only few items~\cite{ColdStartSIGIR02} and sometimes misclick some unwanted items~\cite{NoiseRec}, bringing incompleteness and noisy information to the graph. Moreover, the gap between the original graph and the optimal graph for downstream tasks often exists. For example, the node classification performances of most GNNs are hindered in heterophily graphs, since a homophilic graph, where linked nodes often belong to the same class or have similar features, is a vital premise for GNN-based node classification~\cite{H2GNN,CPGNN}.

Recently, to alleviate this issue, graph structure learning (GSL) is proposed to learn an optimal graph structure instead of relying on the raw graph structure. Most GSL methods adopt a joint learning framework where the estimated graph structure and GNN parameters are optimized together by the downstream task. The graph estimation can be achieved by direct parameterization of the adjacency matrix~\cite{LDS,ProGNN}, or metric learning based on original feature or node embeddings~\cite{IDGL,HGSL}. However, two fundamental weaknesses of this joint framework limit the performance and scalability of GSL methods.
(1) \textit{The inconsistency between graph structure learning and application.} The aim of GSL is to refine the adjacency matrix, which can be viewed as a link prediction task on all potential node pairs. However, the aim of downstream application could be quite different, hampering the graph learning process. Take node classification as an example, with the graph learning parameters optimized towards classifying few labeled nodes, most potential edges in graphs are not well trained~\cite{SLAPS}. (2) \textit{The joint optimization process lacks scalability and efficiency.} On the one hand, most GSL methods require huge memory. Since the adjacency matrix is a $|V| \times |V|$ matrix ($|V|$ denotes number of nodes in the graph), its estimation and optimization takes up huge memory. Moreover, the sampling of anchor nodes~\cite{IDGL} doesn't help much since the computation dependency of GNN grows exponentially in each layer due to the neighborhood aggregation scheme~\cite{LADIES}. On the other hand, with $|V|^2$ potential edges in the adjacency matrix, the calculation and learning of adjacency matrix is time consuming especially on iterative~\cite{IDGL} or bi-level optimized~\cite{LDS} frameworks. 

In light of these limitations, we propose a novel pretrain-finetune framework which alternatively separates the graph structure learning and GNN parameter training. In the pre-training phase, the generation of graph structure is learned by self-supervised tasks; in the fine-tuning phase, a GNN is initialized by the pre-trained model and further fine-tuned on the refined graph by the downstream task. The advantages of this framework are obvious: (1) Different from the joint learning framework, where graph learning is limited by optimizing for downstream tasks, our graph learning is achieved by exploring the unsupervised information which comprehensively explores the underlying graph structure. (2) The ample unlabeled information inside graphs can be migrated to the fine-tuned GNN and further improves the learning process of downstream application. (3) Since the refined graph is static in the fine-tuning phase, the scalability of fine-tuned GNN is almost the same as the original GNN with no additional graph learning module added, enabling large graph applications. However, this is a non-trivial task since there are several challenges remain unresolved. Above all, the underlying graph generation process is driven by complicated and multifaceted reasons~\cite{AM_GCN,HGSL}. It is a more difficult task for GSL without guidance of supervised information. Moreover, the noise and incompleteness of the observed graph further increase the difficulty of learning a reliable graph structure.

In this paper, we make the first attempt to investigate the pretrain-finetune framework for graph structure learning, and propose a novel model \textbf{G}raph \textbf{S}tructure \textbf{R}efinement (GSR). In GSR, a self-supervised pre-training strategy is proposed to explore the underlying mechanism for graph generation. Specifically, multi-view GNNs are trained to perform intra-view and inter-view contrastive learning for link prediction and further refine the original graph. Then, the GNN for downstream task is initialized by the pre-trained GNN and fine-tuned on the refined graph with supervised information. 
The major contributions of this work are highlighted as follows:

\vspace{-0.1in}
\begin{itemize}[leftmargin=*]
    \item To our best knowledge, our work is the first study to investigate the pretrain-finetune framework for graph structure learning. Compared with the mostly adopted joint learning framework, this framework resolves the inconsistency between goals of GSL and downstream task, enabling more scalable GSL.
    \item We propose a novel model GSR which comprehensively explores the underlying mechanism of graph generation by multi-view intra- and inter-contrastive learning. The pre-training process further boosts the fine-tuning process by providing refined graph structure and migrating the pre-trained knowledge.
    \item Extensive experiments are conducted on six benchmark datasets. Notably, GSR outperforms baselines with much less time and space requirement, enabling GSL on large graphs. 
\end{itemize}

\vspace{-5pt}
\section{Related Work}

\subsection{Graph Structure Learning} 
Graph neural networks (GNNs) have drawn significant attention due to their remarkable performance on graph applications~\cite{GCN,ChebNet,GAT,GIN,GNNSurvey,fan2018gotcha,zhang2021rxnet,ye2019out,zhang2019key,ding2022data}. Despite these achievements, most GNNs treat the observed graph structure as ground-truth which significantly hampers their ability to handle uncertainty in graph structure~\cite{BGCN,GNNGuard,AdvAttackOnGraph}. To mitigate this, considerable works have focused on the topic of Graph Structure Learning (GSL), which aims to jointly learn an optimized graph structure and the GNN parameters. The core of existing GSL methods is to estimate an adjacency matrix by a graph learning model and optimize it along with GNN parameters toward the downstream task (e.g. node classification). 
A straightforward design of graph learning module is to directly parameterize the adjacency matrix. For example, TO-GCN~\cite{ToGCN} refines the parameterized adjacency matrix and learns the GNN parameters by class labels. ProGNN~\cite{ProGNN} jointly learns GNN parameters and a robust adjacency matrix with graph properties. Since the direct parameterization of adjacency matrix is time and space consuming, an alternative way is to perform graph learning through metric learning~\cite{AM_GCN,HGSL,GNNGuard,IDGL,SLAPS}. To illustrate, IDGL~\cite{IDGL} iteratively learns the metrics to generate graph structure from node features and GNN embeddings. The estimation of adjacency matrix can also be achieved by probabilistic graph learning models~\cite{BGCN,LDS,GIB,DGM}. The representative one is LDS~\cite{LDS} which samples graph structure from a Bernoulli distribution of adjacency matrix and learns them together with GNN parameters in a bi-level fashion. 
While most GSL works jointly perform graph learning and GNN optimization, which hinders the graph learning process~\cite{SLAPS} and suffers from scalability issues~\cite{GSLSurvey}, our work alternatively proposes a pretrain-finetune framework to alleviate these problems. Besides, Our work is also related to GAuG~\cite{GAuG}, which modifies the graph structure by adding and removing edges based on similarity. Different from GAuG, our work performs GSL in a multi-view contrastive learning framework and enhances the fine-tuning GNN by pre-trained model weights.

\vspace{-0.1in}
\subsection{Self-Supervised Learning on GNNs}
Our work is also related to self-supervised learning (SSL), which leverages the unlabeled input data by unsupervised tasks and benefits a series of applications~\cite{SSLSurvey}. Existing SSL models for GNNs can be categorized into generative or contrastive methods based on their SSL tasks~\cite{SSLSurveyGorC}. On the one hand, generative models learn graph embeddings by recovering feature or structural information on graph. The task can be recovering adjacency matrix alone~\cite{GAE,GraphRNN} or along with the node features~\cite{GPT-GNN}. On the other hand, contrastive models firstly defines the context of a node, which could be node-level, subgraph-level, or graph-level graph instances. Then, contrastive learning is performed by either maximizing the mutual information between the node-context pairs~\cite{MVGRL,DGI,InfoGraph} or by discriminating context instances~\cite{GCC,PretrainGNN,MVSE,yu2022sail}.
Inspired by recent work SLAPS~\cite{SLAPS}, which enhances graph learning by feature denoising on optimized graphs, we utilize SSL in the pre-training phase. Different from SLAPS, which also applies the joint learning framework, we design a pretrain-finetune pipeline. Also, the SSL task of SLAPS and the graph structure refinement process of our model differ largely.

\begin{figure*}[t]\centering
\includegraphics[width=0.95\linewidth]{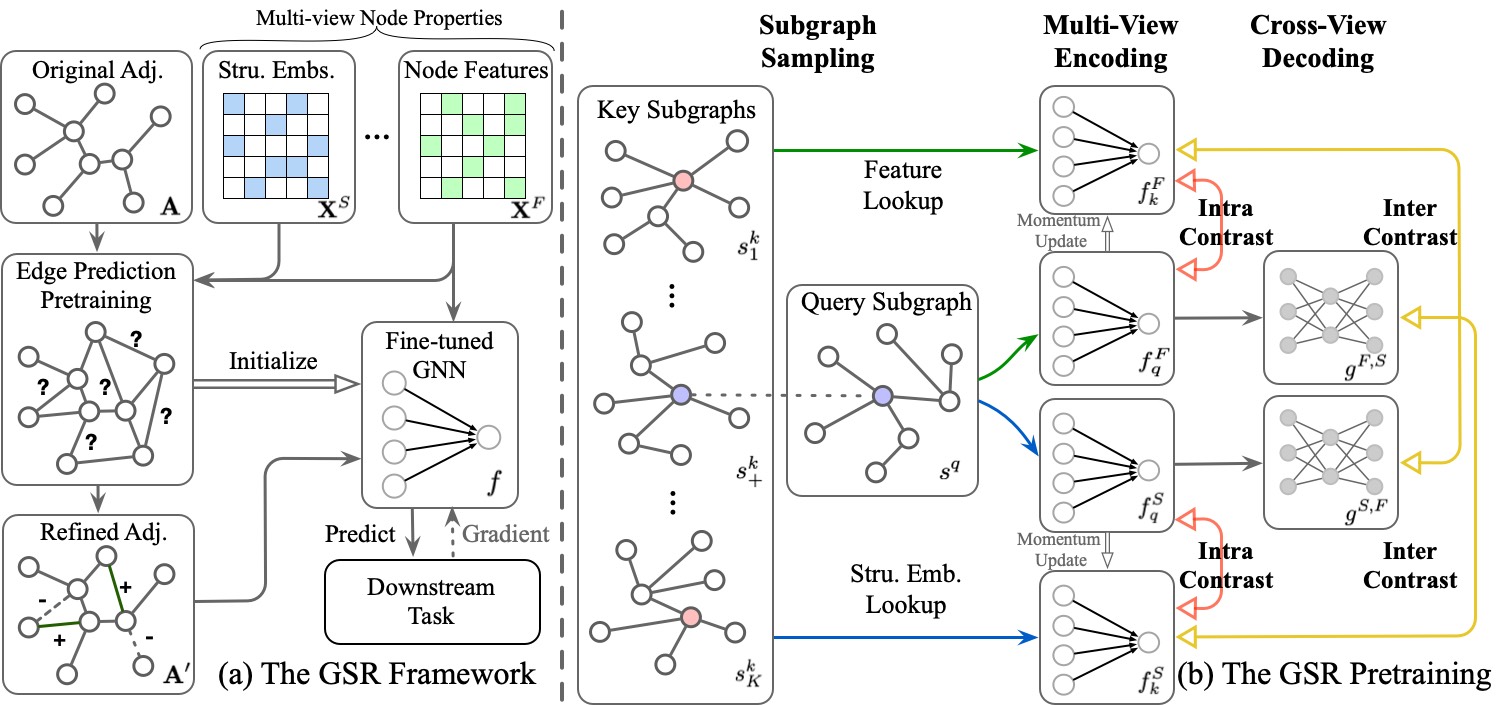}
\vspace{-0.1in}
\caption{Overview of the GSR framework. (a) The proposed GSR framework. (b) The GSR pre-training process. Only two views of node properties, i.e. feature and structural embeddings, are selected for better visualization. Our model can be easily extended to cases with more than two views.}
\vspace{-0.2in}
\label{fig:GSR_model_framework}
\end{figure*}

\vspace{-0.1in}
\section{The Proposed Model}
Figure \ref{fig:GSR_model_framework} (a) illustrates the proposed GSR framework. For a target graph application, GSR solves it through a pretrain-finetune process. In the pre-training phase, GSR learns the formation of observed graph structure by performing link prediction pre-training task on existing edges in a self-supervised manner. Specifically, as the underlying mechanism of the graph generation is multifaceted, GSR encodes multi-view information (e.g., node features, structural embeddings, or other node property specified by users) to predict the probability of each edge and further refine the adjacency matrix. In the fine-tuning phase, the GNN for downstream task is initialized by the pre-trained GNN to migrate the knowledge learned from unlabeled data. Then, the GNN is fine-tuned on the refined graph and optimized by the downstream task.

\vspace{-0.1in}
\subsection{Pretraining with Link Prediction}
Given an observed graph $G=(V,E)$ composed of a node set $V$, an edge set $E$, its adjacency matrix $\mathbf{A}$, and multi-view node property matrices $\mathcal{X}=\{\mathbf{X}^\phi, \phi \in \Phi\}, \mathbf{X}^{F}\in \mathcal{X}$ ($\Phi$ is the set of all views, $\mathbf{X}^F$ is the node feature), the goal of pre-training phase is to train the graph structure learning GNNs that estimate the probability of an edge between a node pair. To achieve this, we design a multi-view contrastive learning process shown in Figure \ref{fig:GSR_model_framework} (b). Specifically, we first sample nodes and their induced subgraphs for contrastive learning. Then, the subgraphs are encoded by aggregating their multi-view embeddings through GNN encoders of corresponding views. The encoded embeddings are firstly optimized by intra-view contrastive objective to discriminate whether an edge exist between two nodes by comparing the similarity between the encoded node embeddings of each view. What's more, the multi-view embeddings are further decoded to other views and perform inter-view contrastive learning, so that the model learns to predict edges with the correlations between different views. Next, we elaborate the contrastive learning process in details.
\\
\textbf{Intra-View Contrastive Learning.}
The intra-view contrastive learning explores the underlying mechanism of graph generation by each view alone. Say we take the original node feature $\mathbf{X}^F$ and the structural embedding $\mathbf{X}^S$ as two-views of the graph, i.e. $\mathcal{X}=\{\mathbf{X}^\phi,\phi \in \{F, S\}\}$, the model encodes different views of information, i.e. $\mathbf{X}^F$ and $\mathbf{X}^S$, to predict whether the node pairs are connected from feature or structure perspectives. 

Inspired by GCC~\cite{GCC}, we utilize MoCo~\cite{Moco} as the contrastive learning framework which contrasts between query and a dictionary of keys. To perform link prediction, a query-key node pair is reckoned as a positive contrastive pair if there exist an edge in graph, otherwise it is a negative pair. Specifically, for a query node $v^q$ and a dictionary of $K + 1$ key nodes $\{v^k_0, \dots ,v^k_K\}$ with their corresponding induced ego-subgraphs $s^q$ and $\{s^k_0, \dots ,s^k_K\}$, the model looks up a single key node $v^k_+$ that has an edge with query node $v^q$ by encoding their corresponding subgraphs from different views. For each view $\phi \in \Phi$ the link-prediction-based contrastive learning is achieved by minimizing the InfoNCE loss~\cite{InfoNCE}. The overall intra-view contrastive learning loss $\mathcal{L}_{intra}$ is defined as the average loss of all views:
\begin{equation}
    \mathcal{L}_{intra}=\frac{1}{|\Phi|}\sum_{\phi  \in \Phi} - \log \frac{\exp \left(\mathbf{z}_q^{\phi} \cdot \mathbf{z}^{\phi}_{k{+}} / \tau\right)}{\sum_{i=0}^{K} \exp \left(\mathbf{z}_q^{\phi} \cdot  \mathbf{z}_{k_{i}}^{\phi} / \tau\right)},
\end{equation}
where $\tau$ is the temperature hyper-parameter, $\mathbf{z}_q^{\phi}$ and $\mathbf{z}_k^{\phi}$ stands for the embedding of $s^q$ and $s^k$ encoded by view-specific query and key GNN encoders $f^{\phi}_{q}$ and $f^{\phi}_{k}$:
\begin{equation}
\label{Eq:Context_encoder}
\mathbf{z}_{q}^{\phi}=f^{\phi}_{q}(\mathbf{A}_q,\mathbf{X}^{\phi}_q), 
\mathbf{z}_{k}^{\phi}=f^{\phi}_{k}(\mathbf{A}_k,\mathbf{X}^{\phi}_k),
\end{equation}
where $\mathbf{A}_q,\mathbf{A}_k$ and $\mathbf{X}^{\phi}_q,\mathbf{X}^{\phi}_k$ are corresponding adjacency matrices and view-specific node properties of $s^q$ and $s^k$. The GNN encoders $f^{\phi}_{q}$ and $f^{\phi}_{k}$ have the same GNN architecture as that of the fine-tuning GNN $f$ for convenience in migrating pre-trained knowledge. The parameters of $f^{\phi}_{k}$, denoted as $\theta_{q}^{\phi}$, are updated by the momentum-based moving average of the parameters of $f^{\phi}_{q}$, i.e. $\theta_{k}^{\phi}$, ~\cite{Moco}:
\begin{equation}
\theta_{k}^{\phi} \leftarrow m \theta_{k}^{\phi} +(1-m) \theta_{q}^{\phi},
\end{equation}
where $m \in[0,1)$ is the momentum coefficient. 
\\
\textbf{Inter-View Contrastive Learning.}
As the observed graph structure and features could be noisy and incomplete~\cite{LDS,IDGL}, the graph structure learned by each view alone may also be affected by these factors. In light of this, we propose to enhance the graph structure learning process by inter-view contrastive learning. Specifically, each view $\phi$ is treated as source view $\phi_s$, and its encoded view embedding $\mathbf{z}_q^{\phi_s}$ of query node $v_q$ is decoded to target view embedding:
\begin{equation}
    \mathbf{\hat{h}}_q^{s, t} = g^{s,t} (\mathbf{z}_q^{\phi_{s}}),
\end{equation}
where $g^{s,t} ( \cdot )$ stands for the decoder (e.g. $\operatorname{MLP}$) that decodes the information from source view $\phi_s$ to target view $\phi_t$. $\mathbf{\hat{h}}_i^{s, t}$ stands for the decode embedding of target view $\phi_t$. Then, the decoded query embedding $\mathbf{\hat{h}}_q^{s, t} $ is further contrasted with target view key embeddings $\mathbf{z}^{\phi_t}_{k}$ to perform inter-view link-prediction-based contrastive learning:

\begin{equation}
{
    \mathcal{L}_{inter}=\frac{1}{|\Phi|(|\Phi|-1)}
    \sum_{\phi_s,\phi_t\in \Phi, s\neq t}-\log \frac{\exp \left(\mathbf{\hat{h}}_q^{s, t} \cdot \mathbf{z}^{\phi_t}_{k{+}} / \tau\right)}{\sum_{i=0}^{K} \exp \left(\mathbf{\hat{h}}_q^{s, t}\cdot  \mathbf{z}^{\phi_t}_{k_{i}} / \tau\right)}.
}
\end{equation}

By optimizing the inter-view contrastive learning loss $\mathcal{L}_{inter}$, GSR enjoys more robust graph structure learning. Take feature and structure views as example, the feature and structure encoders are forced to extract information for structure and feature correlations to perform cross-view link prediction tasks afterwards. Hence, if one view of node property, say feature information, of a node pair is missing or corrupted, the negative impact could be alleviated by its structural information. In this way, more robust pre-training and graph structure estimation are achieved.

Finally, the model optimizes the overall pre-training loss $\mathcal{L}_P$ to comprehensively learn representations considering both intra-view and inter-view contrastive objectives:
\begin{equation}
    \mathcal{L}_P = \alpha\mathcal{L}_{intra} + (1-\alpha)\mathcal{L}_{inter},
\label{eq: L_P}
\end{equation}
where $\alpha$ is the hyper-parameter for balancing two losses.

\vspace{-0.1in}
\subsection{The Fine-tuning Phase}

Once the pre-training phase is completed, we are able to boost the fine-tuning phase in two ways. On the one hand, the graph structure is refined by the pre-trained model. On the other hand, the pre-trained knowledge is migrated to fine-tune model. To start with, the model firstly refines the graph structure and generates the new adjacency matrix $\mathbf{A}^\prime$. Specifically, for a node pair $v_i$ and $v_j$, GSR estimate the edge probability $\mathbf{E}_{i,j}$ using the pre-trained model:
\begin{equation}
\label{eq: Eij}
    \mathbf{E}_{i,j} = \sum_{\phi \in \Phi}\beta_{\phi}\operatorname{Norm}_{\phi} \left(\operatorname{cos}(\mathbf{z}_i^\phi,\mathbf{z}_j^\phi)\right),
\end{equation}
where $\operatorname{cos}(\cdot,\cdot)$ is the cosine similarity function and $\operatorname{Norm}_{\phi}(\cdot)$ stands for a normalization function for probabilities of view $\phi$, $\beta_{\phi}$ is the hyper-parameter to balance the importance of different views. Then, inspired by GAuG~\cite{GAuG}, we refine the graph structure by adding top $m^+$ non-edges with highest probabilities and removing the $m^-$ existing edges with least probabilities. In this way, the raw graph structure is augmented and purified by the pre-trained model.

Though unaware of what the downstream task is, the ample information inside graph data is encoded into pre-trained model by self-supervised learning, making it a good initialization for the fine-tuning GNN~\cite{SSLSurvey,PretrainGNN}. Here, we use the query GNN encoder of node feature view to initialize the fine-tuning GNN. In this way, the knowledge of aggregating feature to perform intra- and inter-view link prediction is transferred. 
Finally, the initialized GNN is fine-tuned on the downstream task using the refined graph. Take node classification as an example, the fine-tuning loss $\mathcal{L}_F$ is thereby defined as:
\begin{equation}
    \mathcal{L}_F=\sum_{v_{i} \in V_{L}} \ell\left(f(\mathbf{\mathbf{A}^\prime},\mathbf{X}^{F})_{i}, y_{i}\right),
\end{equation}
where $f(\mathbf{\mathbf{A}^\prime},\mathbf{X}^{F})_{i}$ is the predicted label of node $v_i \in V_L$ by fine-tuned GNN, $\ell(\cdot, \cdot)$ measures the difference between prediction and the true label $y_i$.

\vspace{-0.05in}
\section{Experiments}

\begin{table*}[ht] 
\centering
\caption{The statistics of the datasets.}
\vspace{-0.1in}
\label{tab: datasets}
\begin{tabular}{|c|ccccccc|}
\hline
Benchmarks  & \#Nodes & \#Edges & \#Classes & \#Features & \#Train      &\#Validation     &\#Test \\ \hline
Cora        & 2,708    & 5,278    & 7         & 1,433       & 140          & 500               &1,000    \\ 
Citeseer    & 3,327    & 4,552    & 6         & 3,703       & 120          & 500               &1,000    \\
Air-USA     & 1,190    & 13,599   & 4         & 238        & 119          & 238               &833    \\
BlogCatalog & 5,196    & 171,743  & 6         & 8,189       & 519          & 1,039              &3,638     \\
Flickr      & 7,575    & 239,738  & 9         & 12,047      & 757          & 1,515              &5,303     \\
ogbn-arxiv  & 169,343  & 1,157,799 & 40        & 128        & 90,941        & 29,799             &48,603    \\ \hline
\end{tabular}
\vspace{-0.1in}
\end{table*}

\begin{table*}[h]
\caption{The accuracy in percentage (mean±std, the best results are bolded) of node classification with standard splits. OOM indicates ``out of memory'' during training process.}
\vspace{-0.1in}
\begin{tabular}{|c|cccccc|}
\hline
Model       & Cora       & Citeseer   & Air -USA   & Blogcatalog & Flickr     & OGB-Arxiv  \\ \hline
GCN~\cite{GCN}        & 81.32±0.59 & 72.48±0.40 & 56.90±0.43 & 76.03±0.19  & 58.25±0.78 & 68.72±0.74 \\ 
GAT~\cite{GAT}        & 82.00±0.62 & 71.50±0.35 & 45.91±1.81 & 64.79±3.45  & 45.28±1.71 & 69.28±0.45 \\ 
GSAGE~\cite{GraphSAGE}  & 82.74±0.73 & 71.40±0.97 & 55.89±1.39 & 73.50±0.50  & 54.75±0.28 & 70.15±0.59 \\ 
SGC~\cite{SGC}        & 80.86±0.24 & 72.20±0.67 & 55.27±0.11 & 74.11±0.60  & 50.97±0.05 & 66.28±0.18 \\ \hline
LDS~\cite{LDS}        & 82.52±0.36 & 72.55±0.32 & 57.63±1.01 & 75.69±0.65  & 60.24±0.72 & OOM        \\ 
IDGL-Anchor~\cite{IDGL} & 83.50±0.20 & 71.20±0.30 & 57.39±0.53 & 75.62±1.17  & 54.29±1.33 & OOM         \\
BGCN~\cite{BGCN}       & 82.64±0.21 & 72.18±0.42 & 56.62±0.88 & 73.97±0.71  & 58.14±1.36 & OOM          \\ 
HGSL~\cite{HGSL}       & 82.66±2.09 & 72.42±1.19 & 60.57±1.58 & 75.92±1.10  & 61.35±1.21 & OOM          \\ 
ProGNN~\cite{ProGNN}     & 81.90±0.32 & 72.94±0.57 & 56.58±1.12 & 74.25±0.50  & 58.38±0.32 & OOM         \\ \hline
GSR (Ours)        & \textbf{83.83±0.80} & \textbf{73.77±0.35} & \textbf{61.58±0.48} & \textbf{76.91±0.22}  & \textbf{62.02±0.25} & \textbf{71.07±0.58} \\ \hline
\end{tabular}
\label{tab:result-1}
\vspace{-0.1in}
\end{table*}
We conduct extensive experiments to comprehensively evaluate the GSR framework. Following most existing GSL works~\cite{LDS,ProGNN,HGSL,IDGL}, we use semi-supervised node classification to evaluate the effectiveness of GSR compared with state-of-the-art GSL methods. Then, we further show that GSR enjoys excellent performance with better scalability and efficiency compared with existing GSL frameworks. What's more, extensive study including ablation study and parameter analysis are performed to evaluate the effectiveness of the model designs. 


\vspace{-0.1in}
\subsection{Experimental Setup}
\subsubsection{Datasets} 
To comprehensively evaluate the effectiveness of GSR, we conduct experiments on the six benchmark graph datasets ranging from citation graphs, social networks, and a air-traffic dataset. The dataset statistics are reported in Table~\ref{tab: datasets}.
\begin{itemize}[leftmargin=*]
    \item \textbf{Citation graphs} (Cora, Citeseer~\cite{GCN}, and ogbn-arxiv are usually used as benchmarks in most GNN literature~\cite{GCN,GAT}. In these networks, the nodes are papers published in the field of computer science; the features are bag-of-word vectors of the corresponding paper title; the edges are the citation relationships between papers; the labels are the categories of papers. 
    \item \textbf{Social networks} (BlogCatalog and Flickr~\cite{BlogFlickr}) are composed of the interactions between online community users. Specifically, BlogCatalog is an online blogging community, where an edge represents the following/followed-by relationship between two users. The features for each user are generated by the keywords in each blogger's description and the labels are selected from predefined categories of blogger interests. Flickr is an image and video-sharing platform composed of users' following/followed-by relationships. The features of each user are extracted from the list of interests tags specified by the user and the groups that users joined are used as labels. 
    \item \textbf{Air traffic graph} (Air-USA~\cite{airport}) is formed by the commercial flights between the airports and Air-USA represents the flights network in the USA. The node labels are generated based on the activity measured by people and flights passed the airports \cite{airport} and one-hot degree vectors are used as node features. 
\end{itemize}


\subsubsection{Baselines} To evaluate the effectiveness of GSR, we compare it with nine baseline methods. We use four GNN baselines including two spectral GNNs, i.e. GCN~\cite{GCN} and SGC~\cite{SGC}, and two spatial GNNs, i.e. GAT~\cite{GAT} and GSAGE~\cite{GraphSAGE}. For GSL baselines, as GSL models can be categorized into three categories based on the graph learning module~\cite{GSLSurvey}, we choose representatives for each type. ProGNN~\cite{ProGNN} is chosen as the representative of models performing graph learning based on direct-optimization of parameterized adjacency matrix. LDS~\cite{LDS} and BGCN~\cite{BGCN} are chosen from probabilistic GSL models. For metric learning based GSL models, GAuG-O~\cite{GAuG}, IDGL-Anchor~\cite{IDGL} and HGSL~\cite{HGSL} are selected.

\begin{figure*}[h]
\centering
\includegraphics[width=1.0\linewidth]{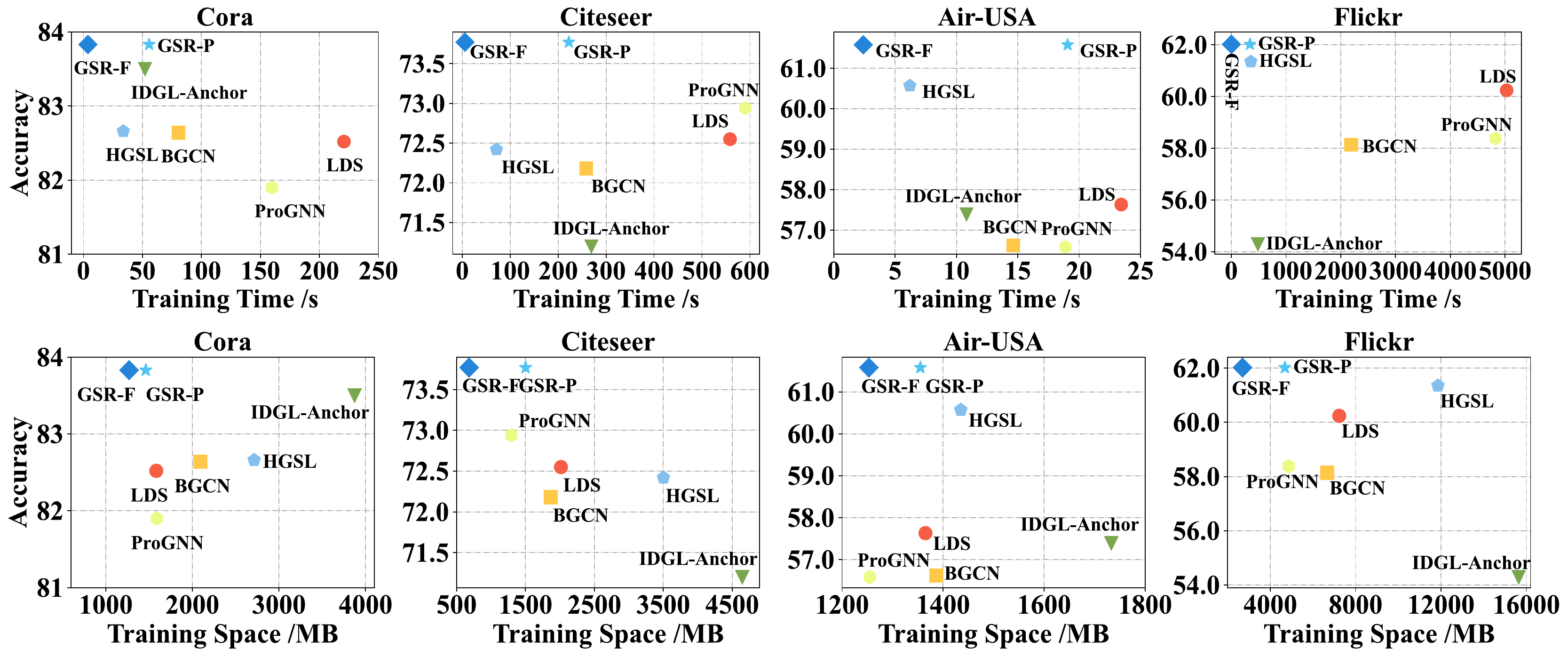}
\vspace{-0.3in}
\caption{Training time and sapce analysis on Cora Citeseer, Air-USA, and Flickr.}
\vspace{-0.2in}
\label{fig:train_time}
\end{figure*}

\subsubsection{Implementation Details} GSR is capable of handling multiple views of node properties to learn the graph structure. We use the node feature and DeepWalk~\cite{DeepWalk} structural embedding as node properties. Therefore, the pre-training model is the same as the model illustrated in Figure \ref{fig:GSR_model_framework} (b). Following previous GSL works, the backbone GNN of GSL models is set as GCN~\cite{GCN}. 

We implement the proposed GSR model with Python (3.8.5) and Pytorch (1.8.1). We provide efficient and scalable code based on DGL~\cite{DGL} (0.6.1). The code and data (available in the supplementary materials) will be made public upon paper publication. For models with DGL~\cite{DGL} implementation, i.e. GCN, SGC, GAT, and GraphSAGE, we use the official DGL implementation to conduct the experiments. The code is available at this github repo\footnote{https://github.com/andyjzhao/WSDM23-GSR}. 

\subsection{Node Classification Performance}
\subsubsection{Results on Standard Splits}
 The results of all models with standard train-validation-test split are shown in Table \ref{tab:result-1}, from which we have the following observations: 
(1) With the ability to learn graph structure, GSL based methods generally outperform GCN, indicating that the raw graph structure is not optimal for node classification and further demonstrates the necessity of graph structure refinement. 
(2) GSR consistently outperforms all the baselines on all six datasets, which proves the effectiveness of our proposed model. Notably, even the graph structure learning process of GSR is unaware of downstream task, the SSL refined graph structure still boosts the performance of node classification. (3) With the joint learning framework, all GSL baselines can not be trained due to OOM errors on OGB-Arxiv (169,343 nodes). On the contrast, with the pretrain-finetune framework, GSR is able to boost GCN on OGB-Arxiv with only 3.1GB GPU memory.
\subsubsection{Results on Different Label Rate.}
With the ability of utilizing rich unlabeled data, one great advantage of SSL lies in the capability of dealing cases when supervision is scarce~\cite{SSLSurvey}. Here, we verify whether GSR also has this ability by conducting experiments with different amount of supervised information. We conduct experiments on stratified splitted datasets (Cora and Citeseer) with training ratio 1\%, 3\%, 5\% and 10\%. After the training nodes are selected according to training ratio, the rest of labeled nodes are evenly splitted into validation set and testing set. The results are shown in Table \ref{tab: label_rate}, where we find GSR outperforms the other baselines in most of the scenarios. Notably, on Citeseer, GSR achieves large improvements over other baselines when the supervised information is limited, which demonstrates the effectiveness to fully utilize the unsupervised information in the pre-training phase. Apart from GSR, HGSL also achieves better performance than most of the other baselines. The effectiveness of GSR and HGSL imply that it is better to refine the original graph structure based on multi-view (feature and structure) knowledge.

\vspace{-0.1in}
\subsection{Efficiency and Scalability Analysis}

\begin{table*}[ht]
\caption{The accuracy in percentage (mean±std, the best results are bolded) of node classification with different train ratios.}
\vspace{-0.15in}
\begin{tabular}{|c|c|c|c|c|c|c|c|c|}
\hline
\multirow{2}{*}{Model} & \multicolumn{4}{c|}{Cora}                         & \multicolumn{4}{c|}{Citeseer}                      \\ \cline{2-9} 
                       & 1\%        & 3\%        & 5\%        & 10\%       & 1\%        & 3\%         & 5\%        & 10\%       \\ \hline
GCN                    & 59.31±0.29 & 77.14±0.21 & 80.73±0.63 & 83.53±0.42 & 60.64±1.07 & 67.60±0.47  & 70.05±0.54 & 74.38±0.27 \\ 
GAT                    & 65.36±0.99 & 76.36±0.61 & 81.73±0.21 & 83.92±0.42 & 58.48±2.35 & 68.41±0.76  & 70.73±0.22 & 74.54±0.14 \\ 
GraphSage              & 68.04±0.81 & 77.29±1.13 & 82.80±0.67 & 84.84±0.49 & 58.97±1.12 & 66.93±0.15  & 70.12±0.77 & 74.35±0.51 \\ 
GIN                    & 55.57±3.99 & 69.64±1.53 & 75.95±0.94 & 79.70±1.10 & 48.29±2.46 & 58.92±1.73  & 61.67±1.96 & 67.22±1.78 \\ 
LDS                    & 68.47±1.38 & 78.06±1.07 & 81.42±0.69 & 83.87±0.43 & 61.35±1.82 & 67.29±1.24  & 70.82±0.87 & 74.54±0.45 \\ 
IDGL-Anchor            & 70.83±1.16 & 78.60±0.24 & 83.82±0.27 & 85.51±0.09 & 60.61±1.32 & 64.34±14.93 & 69.39±1.25 & 74.19±0.53 \\ 
BGCN                   & 66.21±2.35 & 75.71±1.10 & 79.63±0.87 & 82.94±0.36 & 59.42±2.05 & 66.17±0.85  & 70.37±0.65 & 74.16±0.62 \\ 
HGSL                   & \textbf{72.72±3.40} & 79.54±1.28 & 83.56±2.77 & 84.85±0.53 & 61.76±0.83 & 70.69±0.71  & 71.97±0.47 & 74.33±0.45 \\ 
GAuG-O                 & 65.94±4.62 & 73.37±0.69 & 79.14±0.25 & 82.37±0.19 & 57.85±1.95 & 67.67±0.86  & 70.53±0.14 & 74.32±0.08 \\ 
ProGNN                 & 70.25±1.31 & 75.93±0.67 & 81.35±0.62 & 82.01±0.56 & 56.77±0.94 & 70.34±0.69  & 70.67±0.88 & 74.23±0.39 \\ \hline
GSR                    & 72.33±0.47 & \textbf{79.99±0.20} & \textbf{84.30±0.21} & \textbf{85.81±0.22} & \textbf{66.22±0.49} & \textbf{71.89±0.16} & \textbf{72.38±0.22} & \textbf{75.17±0.20} \\ \hline
\end{tabular}
\vspace{-0.15in}
\label{tab: label_rate}
\end{table*}


Here, we analyse the efficiency and scalability of the GSL models. As discussed in section \ref{sec: intro}, the estimation and optimization of adjacency matrix in the joint learning framework are time and space consuming. The training time and GPU space requirement are shown in Figure \ref{fig:train_time}. We can see that, in terms of training time, LDS and ProGNN takes much longer time to train with $|V|^2$ parameters in the graph learning module. Meanwhile, metric learning based GSL models, i.e. IDGL and HGSL, are faster with less parameters. In terms of space requirement, IDGL-Anchor requires the largest space even the anchor nodes are selected to alleviate scalability issues. Notably, the proposed GSR framework outperforms other GSL models with remarkable efficiency and scalability in terms of both training time and space requirement. Moreover, since pre-trained models are commonly used in an off-the-shelf manner to improve fine-tuning performance, a fast and scalable fine-tuning phase becomes a critical requirement for GSL model in large graph applications. From this angle, the superiority of GSR becomes more obvious as GSR-F is 13.8$\times$ faster using only 32.8\% GPU space compared to the best GSL baseline IDGL-Anchor. We also notice that, the superiority of GSR becomes more obvious when it comes to larger graphs. For example, GSR achieves better performance with 99.42 $\times$ faster and using 22.28\% GPU memory compared to the best GSL baseline HGSL in Flickr. Notably, all GSL baselines require huge space to train and leads to OOM error (on device with 16GB GPU memory) on ogbn-arxiv. Meanwhile, GSR is able to run on ogbn-arxiv with 285 seconds and 58 seconds of pre-training and fine-tuning time with 2155MB and 3325MB GPU memory in the pre-training and fine-tuning phase.





\vspace{-0.1in}
\subsection{Analysis on the Refined Graph Structure}


GSR refines the graph structure by adding non-existing edges and removing existing edges. As discussed in previous works~\cite{H2GNN,CPGNN}, a homophilic graph is an important premise in node classification for GNNs. Thus, we use homophily ratio~\cite{CPGNN, GAuG}, defined as $\frac{|E_{i}|}{|E|}$, where $E_{i}$ denotes the edges whose nodes are of same label, and node classification accuracy to evaluate the quality of learned graphs.

\subsubsection{Evaluation of Adding Non-existing Edges.} 
\begin{figure}[h]
\centering
\includegraphics[width=1\linewidth]{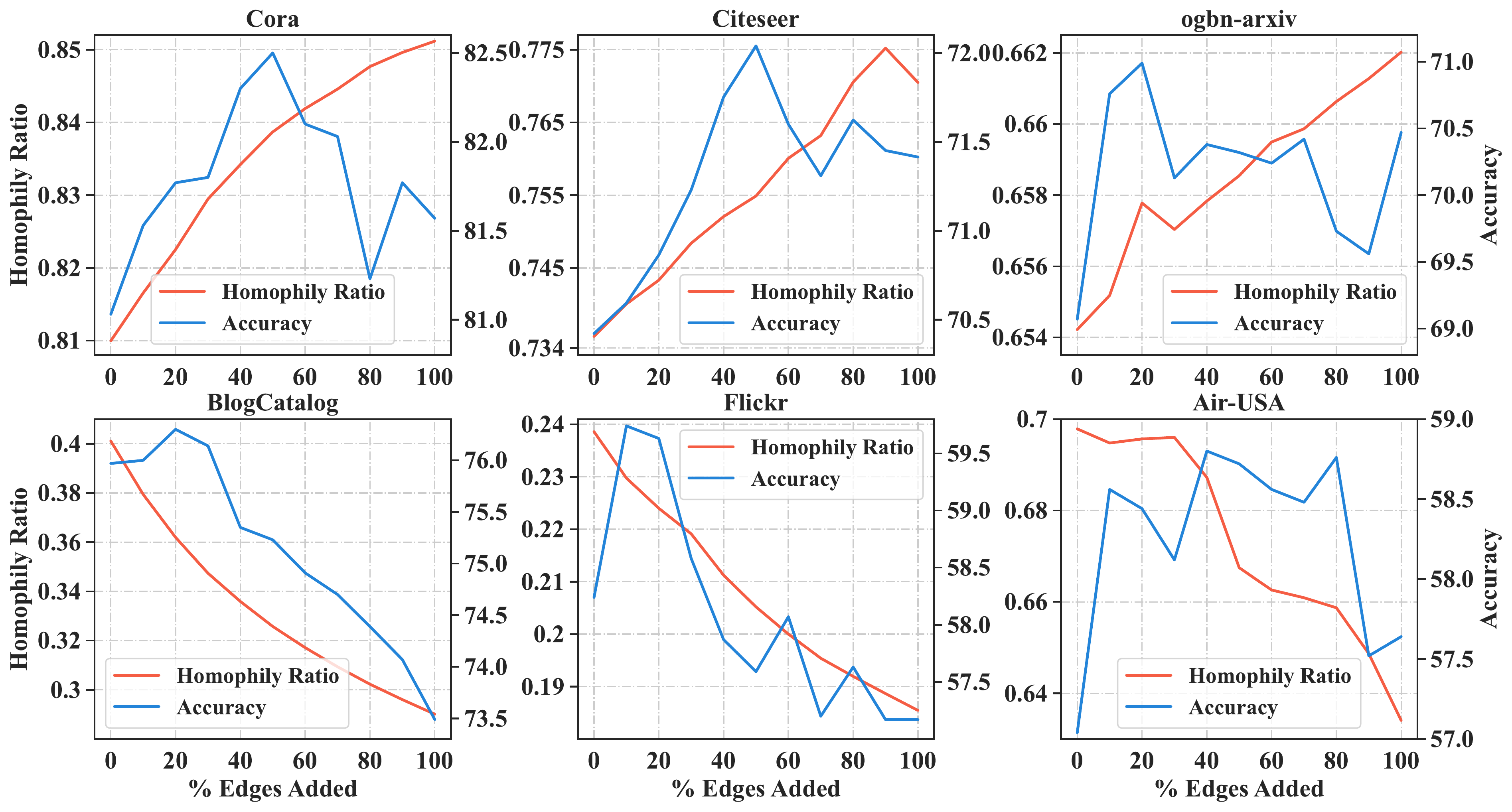}
\vspace{-0.3in}
\caption{Homophily ratio and accuracy in terms of number of added non-existing edges.}
\vspace{-0.2in}
\label{fig:homophily_add}
\end{figure}

From Figure \ref{fig:homophily_add}, we can observe that in all graph datasets, adding appropriate number of edges boosts the performance of node classification. But the performance decreases when too many edges are added. The potential reasons are two-fold: On the one hand, the edges are added in the order of estimated probabilities, adding too much edges with little confidence adds noise to the graph. On the other hand, the message-passing scheme of vanilla GCN~\cite{GCN}, where aggregation is performed with equal edge weights. Therefore, adding too many edges decreases the weights of original edges (most of which are reliable) and harms the performance. Additionally, the learned graph structure is not guaranteed to be homophilic: For citation graphs, i.e. Cora, Citeseer, and ogbn-arxiv, the homophily ratio improves with more learned edges added. However, in social network graphs (BlogCatalog and Flickr) and Air-USA, the homophily ratio decreases. This is probably due to the inconsistency between the node properties for learning graph structure and node classification objective. For example, in BlogCatalog, a user's description (node feature) may not reflect the user's interest (label). 

\subsubsection{Evaluation of Removing Existing Edges.} 
\begin{figure}[h]
\centering
\vspace{-0.1in}
\includegraphics[width=1\linewidth]{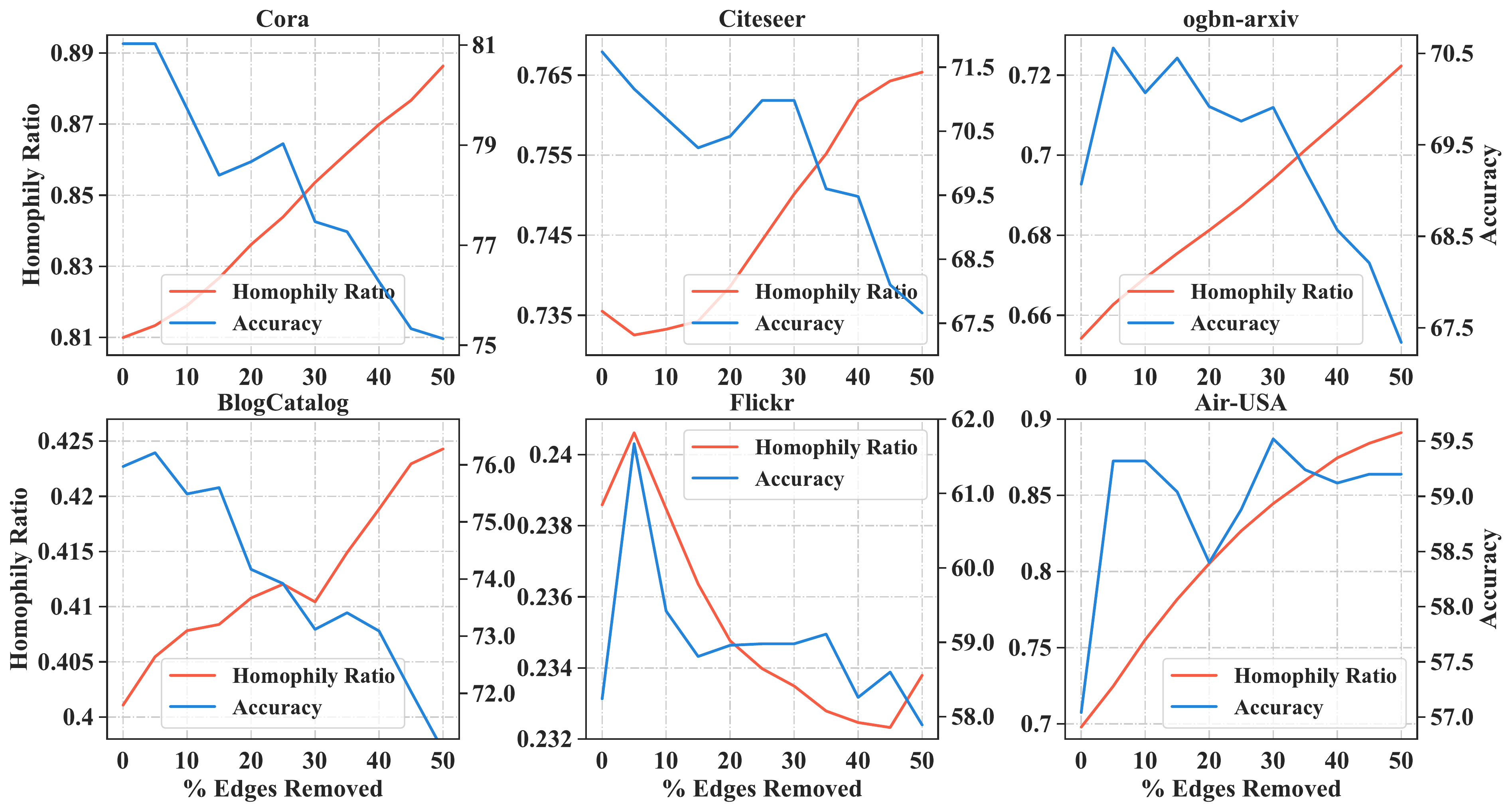}
\vspace{-0.3in}
\caption{Homophily ratio and accuracy in terms of number of removing existing edges.}
\vspace{-0.2in}
\label{fig:homophily_remove}
\end{figure}

We also evaluate the impact of removing existing edges in Figure \ref{fig:homophily_remove}. We can observe that removing a small fraction of edges with low estimated existing probability generally helps constructing a more homophilic graph and boosts the performance of node classification in most cases, which indicates that most edges in the original graph structure are reliable and should not be removed. The phenomenon becomes more obvious in citation datasets such as Cora and Citeseer, where original edges (citation between papers) are mostly correct and removing edges harms the node classification performance.

\begin{figure}[h]
\centering
\includegraphics[width=1.0\linewidth]{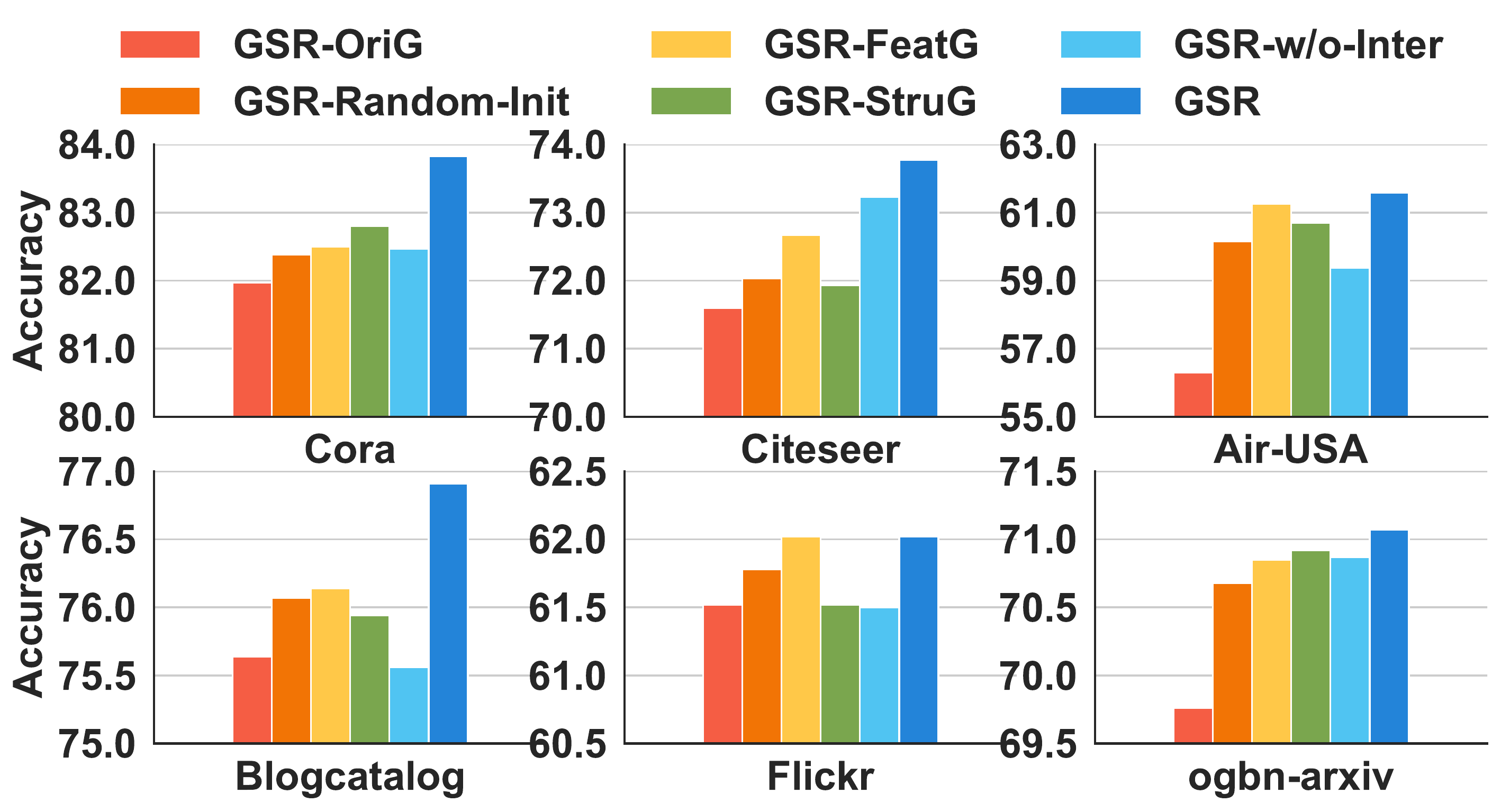}
\vspace{-0.25in}
\caption{The performance evaluation of variants of GSR.}
\vspace{-0.25in}
\label{fig:ablation_study}
\end{figure}

\vspace{-0.1in}
\subsection{Ablation Studies}
To verify the effectiveness of different modules in GSR, we design five GSR variants and evaluate their performances in Figure \ref{fig:ablation_study}.

\subsubsection{Evaluation of SSL Modules}
The proposed GSR pre-training module trains GNN encoders by multi-view contrastive link prediction tasks. Since the intra-contrastive learning part can be roughly viewed as a multi-view ablation of unsupervised GraphSAGE~\cite{GraphSAGE}, which is well studied to be effective in encoding graph structure data, we focus on verifying whether inter-view contrastive learning helps. Therefore, we propose GSR-w/o-Inter, in which the inter-view contrastive learning part is removed. As observed, the significant performance decrease of GSR-w/o-Inter on all datasets compared to GSR demonstrates the importance of inter-contrastive learning.
We further evaluate whether the knowledge learned from unlabelled data is beneficial in the fine-tuning phase. We design GSR-Random-Init, in which the fine-tuned GNN is randomly initialized and trained from scratch. We can see that GSR generally outperforms GSR-Random-Init, demonstrating that the merits of migrating the pre-trained knowledge.

\subsubsection{Effectiveness of the Graph Refinement Module}
Here we evaluate the graph refinement module of GSR. Recall that GSR use the pre-trained multi-view link prediction model to refine the graph structure, we aim to verify whether the learned graph is beneficial for downstream application and whether the multi-view graph estimation module is necessary. 
Thus, we design three variants of GSR, namely GSR-OriG, GSR-FeatG, and GSR-StruG, where the fine-tuning GNN is trained on the original graph strucutre, graph structure based on feature estimation only, and graph structure based on structural estimation only. From the results shown in Figure \ref{fig:ablation_study}, we have following findings: The graph refinement process is of vital importance in the GSR framework, as the performance of GSR with the original graph decreases significantly. The graph refinement with multi-view link prediction is more beneficial compared to graph refinement with single-view, given the fact that GSR outperforms all the ablations of graph refinement. Moreover, we also notice that the performance decrease of GSR-FeatG and GSR-StruG compared to GSR varies in different datasets, indicating that the views should be carefully balanced.

\vspace{-0.1in}
\subsection{Parameter Analysis}
\subsubsection{Impact of $m^+$ and $m^-$}
\begin{figure}[t]
\centering
\includegraphics[width=1.0\linewidth]{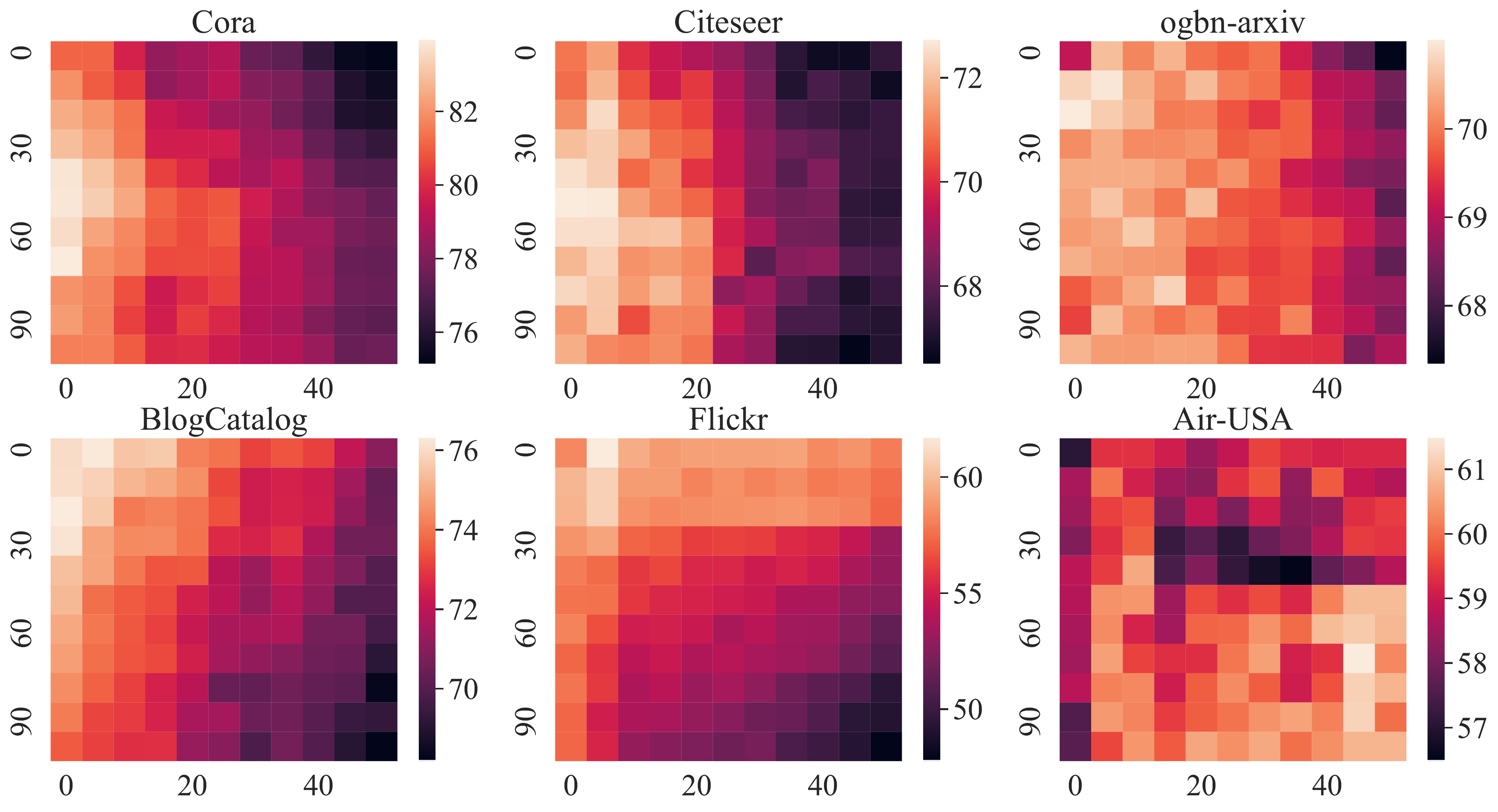}
\vspace{-0.3in}
\caption{The impact of adding $m^+\in [0,|E|]$ and removing edges $m^-\in [0,0.5|E|]$ in the graph refinement process (Cora on the left, Citeseer on the right). Lighter color denotes better node classification performance.}
\vspace{-0.25in}
\label{fig:heat_map}
\end{figure}

The impacts of adding and removing edges are shown in Figure \ref{fig:heat_map}, from which we can observe that: (1) In citation datasets, i.e. Cora, Citeseer and ogbn-arxiv, where the task is to predict the category of each paper, adding edges boosts the performance of GNN, while removing edges doesn't help much in most cases. A probable reason is that these graphs are sparse with little noise, since a paper in a citation graph commonly cites only few most-related papers, leading to a sparse graph with little noise. Hence, adding edges discovers potentially related citation relationships while removing existing edges harms the graph structure. (2) In social network graphs, where the task is to predict the users' interest based on their following/followed-by relationships. Removing edges is of more use than adding edges in the graph refinement process. A probable reason is that the following/followed-by interactions are complex and may not reflect the classification objective (e.g. users' interest in BlogCatalog) sometimes, e.g. a user may follows his friends with different interest, removing edges can be viewed as an effective denoising process that boosts the GNN performance. (3) In Air-USA, where class labels are assigned based on the nodes' degree, adding and removing edges both help. The different impacts of adding and removing edges across datasets demonstrate that the optimal graph structure is related to the characteristics of datasets and the corresponding tasks. Therefore, we should carefully balance the number of adding and removing edges.

\subsubsection{Impact of $\alpha$}
\begin{figure}[h]
\centering
\includegraphics[width=\linewidth]{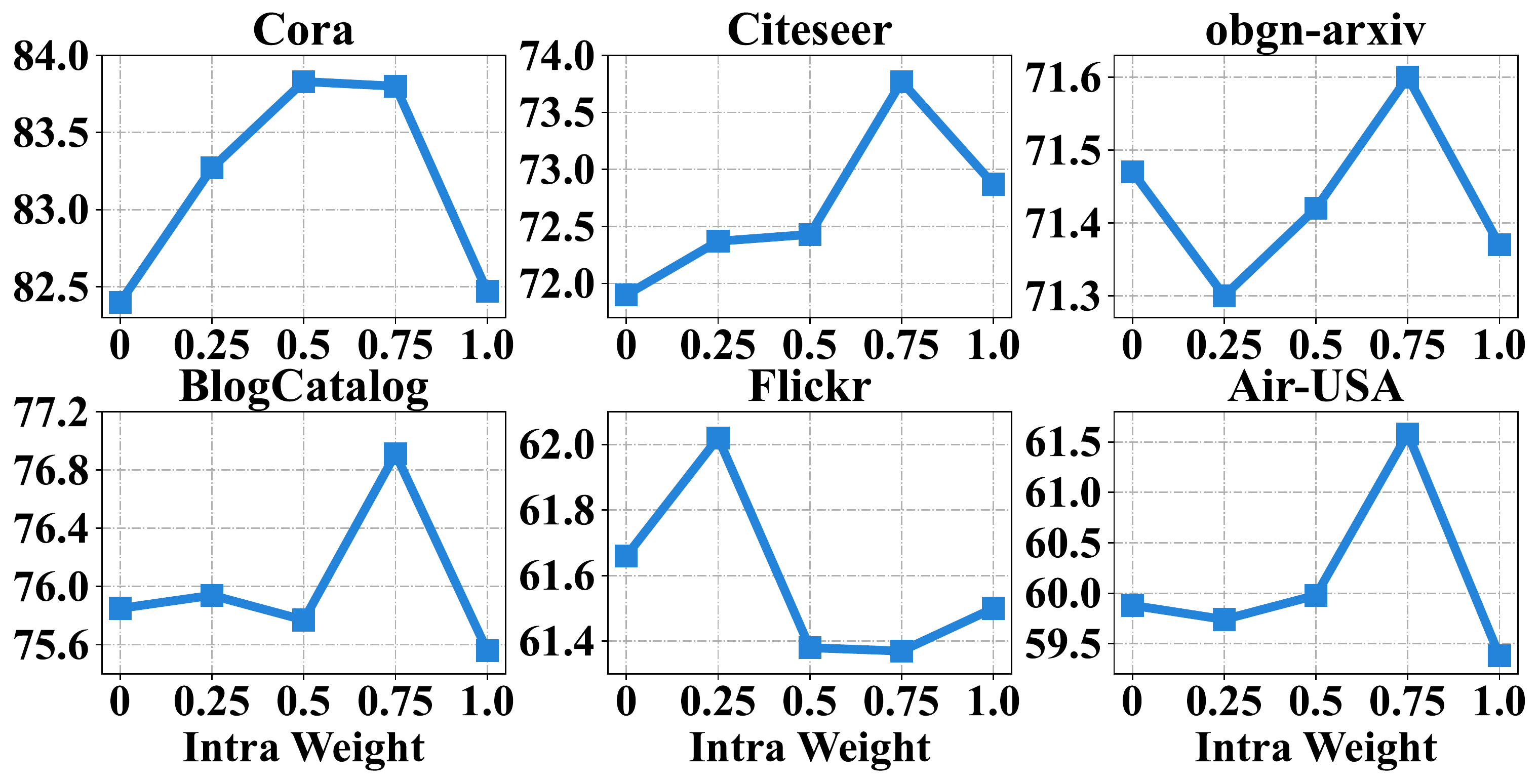}
\vspace{-0.2in}
\caption{Impact of varying the $\alpha$.}
\vspace{-0.2in}
\label{fig:parameter_analysis_alpha}
\end{figure}

In the pre-training phase, GSR uses $\alpha$ to balance the intra- and inter-contrastive loss. As shown in Figure \ref{fig:parameter_analysis_alpha}: The optimal $\alpha$ is around 0.75 in most datasets, which indicates that the intra-contrastive learning is more important. However, it is also important to consider both intra- and inter-loss since the performance decreases when only intra- ($\alpha=1$) and inter-contrastive loss ($\alpha=0$) is concerned. 
\subsubsection{Impact of $\beta$} 
\begin{figure}[h]
\centering
\includegraphics[width=\linewidth]{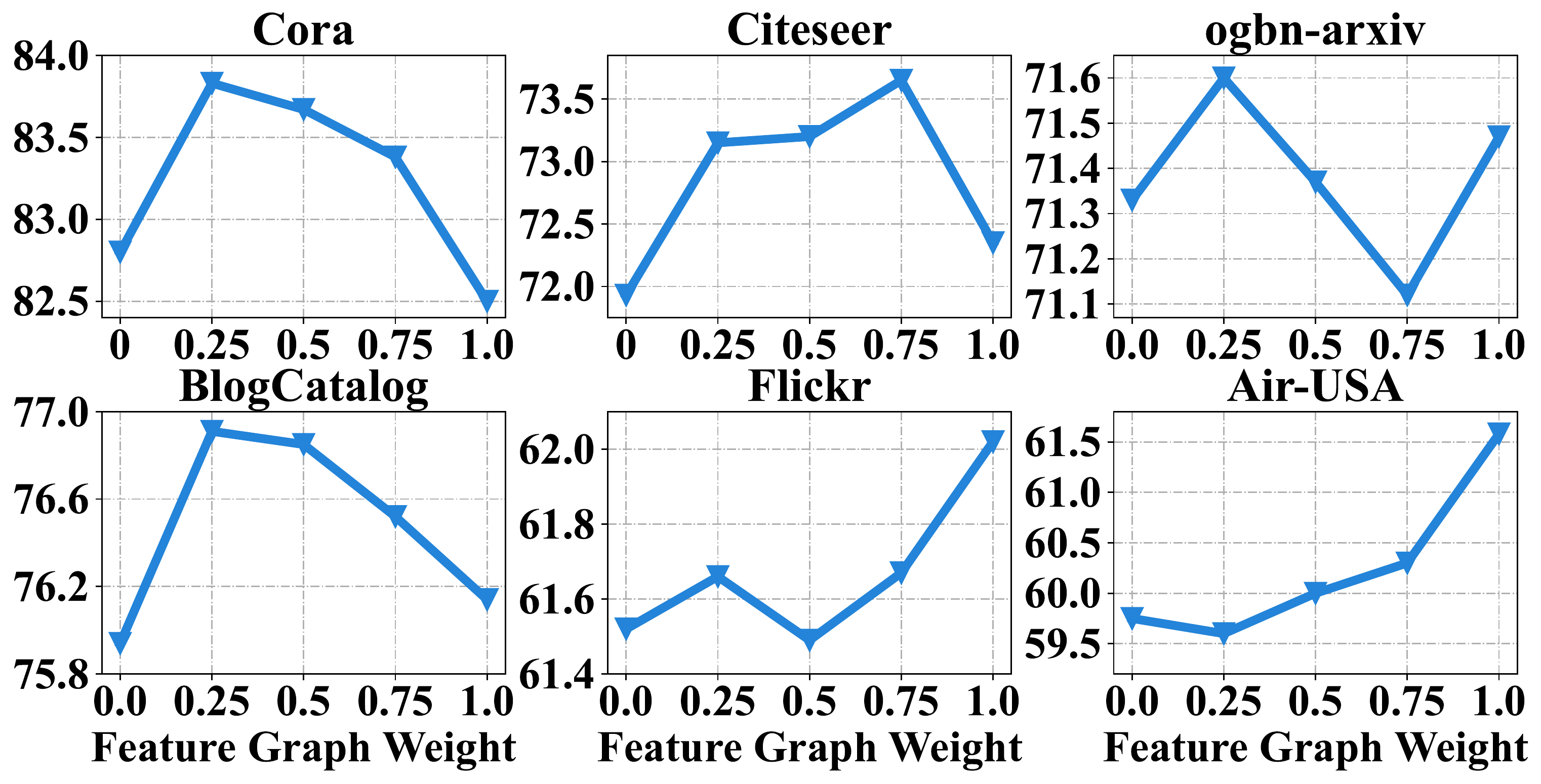}
\vspace{-0.2in}
\caption{Impact of varying the $\beta$.}
\vspace{-0.2in}
\label{fig:parameter_analysis}
\end{figure}

We use two views of information, i.e. feature and structural embedding, in the experiments. Therefore, the graph structure is estimated with feature graph weight $\beta_{F}$ and structure graph weight $\beta_{S}=1-\beta_{F}$. From Figure \ref{fig:parameter_analysis} We can observe that in most datasets, the optimal performance is achieved when feature and structure graph are both considered, demonstrating the effectiveness of utilizing multi-view knowledge on boosting the downstream task performance. Besides, the impact of feature graph weights on performance differs in datasets, indicating that the importance of the graph structure generated by different view is dataset-dependent and should be carefully balanced.

\section{Conclusion}
In this paper, we propose a pretrain-finetune framework GSR for graph structure learning. In GSR, multi-view contrastive link prediction tasks are proposed to fully explore the underlying mechanism for graph generation. The pre-training process further boosts the fine-tuning process by refining graph structure and migrating the pre-trained knowledge. Extensive experiments are conducted to demonstrate the effectiveness, scalability, and efficiency of GSR.
\vspace{-0.2pt}
\section*{Acknowledgments}
This work is partially supported by the NSF under grants IIS-2209814, IIS-2203262, IIS-2214376, IIS-2217239, OAC-2218762, CNS-2203261, CNS-2122631, CMMI-2146076, and the NIJ 2018-75-CX-0032. Any opinions, findings, and conclusions or recommendations expressed in this material are those of the authors and do not necessarily reflect the views of any funding agencies.

\newpage
\bibliographystyle{ACM-Reference-Format}
\bibliography{gsr.bib}

\end{document}